\documentclass{article}
\usepackage{ijcai19}

\usepackage{times}
\usepackage{soul}
\usepackage{url}
\usepackage[hidelinks]{hyperref}
\usepackage[utf8]{inputenc}
\usepackage[small]{caption}
\usepackage{graphicx}
\usepackage{amsmath}
\usepackage{booktabs}
\usepackage{algorithm}
\usepackage{algorithmic}
\usepackage{microtype}
\usepackage{graphicx}
\usepackage{subfigure}
\usepackage{amsmath}
\usepackage{amsthm}
\usepackage{amsfonts}
\usepackage{amssymb}

\usepackage{hyperref}

\newcommand{\E}{\mathrm{E}}

\newtheorem{definition}{Definition}
\newtheorem{proposition}{Proposition}

\author{
Dotan Di Castro$^1$
\and
Joel Oren$^1$\and
Shie Mannor$^{2}$
\affiliations
$^1$Bosch Center for Artificial Intelligence, Haifa, Israel.\\
$^2$Technion – Israel Institute of Technology\\
\emails
\{dotan.dicastro, joel.oren\}@il.bosch.com,
shie@ee.technion.ac.il
}
\title{Practical Risk Measures in Reinforcement Learning}

\begin{document}
\maketitle




\begin{abstract}
Practical application of Reinforcement Learning (RL) often involves risk considerations. We study a generalized approximation scheme for risk measures, based on Monte-Carlo simulations, where the risk measures need not necessarily be \emph{coherent}. We demonstrate that, even in simple problems, measures such as the variance of the reward-to-go do not capture the risk in a satisfactory manner. In addition, we show how a risk measure can be derived from model's realizations. We propose a neural architecture for estimating the risk and suggest the risk critic architecture that can be use to optimize a policy under general risk measures. We conclude our work with experiments that demonstrate the efficacy of our approach.
\end{abstract}

\section{Introduction}
\label{submission}
Reinforcement learning (RL) is a framework where an agent interacts with an unknown environment \cite{sutton2018reinforcement,bertsekas2005dynamic} and the agent can optimize its performance based on a scalar feedback from the environment. The objective in this framework is typically to maximize (or minimize) a cumulative reward (or cost). However, in many cases the agent is tasked with solving the problem at hand while minimizing some form of \emph{risk}, which is a measure that quantifies potential worst case scenarios, that may result from the agent's solution (a \emph{policy}).

There are many examples of risk-aware decision making problems. Important cases include process control (where one is looking for optimizing a process but without endangering the deliverables), finance (where one is looking to avoid catastrophic financial events), motion control (where one, for example, is looking for safety in a shared space where humans and robots are working together), or automotive (where one is interested in safe plan for lane change in self-driving car).

The main vehicle we employ in this work in order to incorporate risk into the optimization of an agent is the \emph{policy gradient} method (PG; \cite{glynn1990likelihood,baxter2001infinite}). As was shown by \cite{tamar2012policy}, extending the PG method for risk is straight forward. We further develop the PG method in this work in order to be able plugin in any risk function.

In this work we show that even the undoubtedly rich class of coherent risk functions\footnote{A risk function $f$ is said to be \textit{coherent} if it satisfies (A1) Convexity: $\forall 0\le \lambda\le 1$, $f(\lambda Z+(1-\lambda) W  \le \lambda f(Z)+(1-\lambda)f( W)$; (A2) Monotonicity: if $Z<W$ then $f(Z)<f(W)$; (A3) Translation Invariance: $\forall \alpha \in \mathbb{R}$, $f(Z+\alpha) = f(Z) + \alpha$; (A4) Positive Homogeneity: if $\lambda \ge 0$, $f(\lambda Z) = \lambda f(Z)$.} \cite{shapiro2009lectures} does not fully capture the desired practical properties of risk functions (for example, look at the non-standard risk functions suggested by \cite{liu2006functional,liu2012existence}). Also, the state-of-the-art methods in coherent risk measure optimization involve complicated machinery \cite{tamar2015policy}, whereas our proposed method provides a significant improvement in terms of computational efficiency, as well its relative simplicity.


The main contributions of this work are the following:
\begin{enumerate}
    \item In our proposed architecture, we allow general risk measures. Although, coherent risk measures (as in \cite{tamar2015policy}) are quite general, still for some use cases one may need other risk shapes.
    \item We show how to implement an RL agent that considers a risk measure with a deep neural architecture \cite{lecun2015deep,goodfellow2016deep}. Since deep networks are powerful function approximators, almost any risk function that one can think off can be plugged into our architecture. Our proposed architecture is based on the Actor Critic method (\cite{sutton2018reinforcement,konda2000actor}). 
    
    Deep Q-Networks (DQN; \cite{mnih2015human}) are used in a \emph{bootstrapping} fashion to estimate the value function by minimizing the square Temporal Difference (TD; \cite{sutton1988learning}). Such methods are broadly ineffective for estimating risk functions, apart from a notable exceptions (e.g., variance). Therefore, our proposed method is based mainly on Monte-Carlo simulations. 
    \item We demonstrate how to shape the risk function. Most of the previous works assume an arbitrary risk function, where its fitness to the problem at hand is unknown. Similarly to reward-shaping, we suggest \emph{risk-shaping} and a simple process to extract from observed data the risk function. The induction for this approach comes from the reward-shaping literature \cite{ng1999policy}, where it was demonstrated that a better shaped reward function can increase an algorithm performance.
\end{enumerate}

We note that Risk Shaping and Generalized utility functions are complementary to each other. As we will show, when extracting from data a risk function, we are not guaranteed that the extracted risk function is coherent. 

The paper is organized as follows. In Section \ref{sec:related_work} we review related work and in Section \ref{sec:setup} we formulate the problem. In  Section \ref{sec:risk_shaping} we discuss risk shaping. In Section \ref{sec:neural_architectures} we provide neural architectures for solving the risk problem. In Section \ref{sec:experiments} we demonstrate our findings. We present our conclusions and future work in Section \ref{sec:conclusions}.

\section{Related Work}
\label{sec:related_work}
The literature of risk in MDPs is quite rich and dates back to \cite{howard1972risk}. \cite{sobel1982variance} was the first to show that the risk measures on the reward-to-go can be written in a closed form, but solving MDPs (both for planning or RL) and incorporating risk measures based on these closed forms is practically impossible due to its high non linearity. Another analytical direction is the \emph{exponential utility function} risk measure \cite{mihatsch2002risk}. Although this form is highly analytical, it hardly captures real world problems.

In recent years, the interest in Risk in MDPs gained a new interest. A very basic form of risk is constraining the instantaneous variance of a state as investigated in \cite{altman1999constrained,geibel2005risk,sato2001td}. This problem is of polynomial complexity and can be solved easily. \cite{mannor2011mean} had shown that if one tries to optimize an MDP where a constraint on the variance of the reward-to-go is given the problem in hand may be NP-Hard and only a local solution for the optimization is possible.

In the context of RL and planning where the risk criteria is the variance of the reward-to-go such local optimal solutions were given explicitly by \cite{tamar2012policy} to the policy gradient method \cite{glynn1990likelihood}. In the context of policy evaluation, \cite{tamar2013temporal} provides a way to incorporate the variance of the reward-to-go for TD and LSTD methods. \cite{prashanth2013actor} suggest an actor critic algorithm.

Another direction of research is of the Value at Risk (VaR) criteria and limiting the percentile \cite{filar1995percentile,rockafellar2000optimization,morimura2012parametric,chow2017risk}. In this case, we want to limit the worst cases trajectories starting from a specific state. Another related method is the Conditional Value at Risk (CVaR; \cite{tamar2015optimizing,chow2017risk}). In this method the objective is to estimate the average of some lower percentile in contrast to the  percentile distribution. For both methods, it seems that one cannot escape estimating the probability distribution of the trajectories starting for a specific state, therefore, making our proposed simplistic architecture a bit more straight forward.

A generalization for both the variance and the CVaR in the context of RL is the \emph{coherent risk} \cite{shapiro2009lectures}. In this generalization, some ``good traits" of a risk function are considered (among them, convexity, insensibility to constants, scaling of risk, etc. \cite{tamar2015optimizing} developed a closed form formula for policy gradient with generalized risk measure in the context of MDPs and RL. Finding  the solution for the coherent risk measures is not an easy task: one needs to solve a constrained optimization problem which may differ for different risk functions. Our approach on the other hand is a ``plug-and-play" approach, where for each desired risk function one can use it in a straight forward manner in our architecture.

Our method of estimating the risk is \emph{Monte-Carlo simulation} \cite{fishman2013monte} and a variant of \emph{Model Predictive Control} (MPC; \cite{camacho2013model}). We use samples of past experiences for estimating the risk of different states.  

Another body of work that relates to Risk is the Constrainted MDPs literature (CMDP; \cite{altman1999constrained}). The CMDP case can be viewed as risk case where the risk function is the special case of the identity function. A variant of constrained MDPs is ``constrained policy iteration" \cite{achiam2017constrained}. In this case, constraints on the policy computation itself are applied. 

The subject of \emph{Safety in Reinforcement Learning} is intimately related to risk in MDPs  \cite{amodei2016concrete,garcia2015comprehensive,ammar2015safe,moldovan2012safe,pirotta2013safe}. In safety, the objective is to satisfy some constraints on the policy whereas in risk the objective is not to violate some measure that quantifies the risk.

\section{Setup and Basic Formulae}
\label{sec:setup}
We consider a Markov Decision Process (MDP; \cite{puterman1994markov}) where $X$ and $U$ are the state space and action space, respectively. The probability  $P(y|x,u)$ is the transition probability from state $x\in X$ when applying action $u \in U$ to the state $y\in X$. For this transition matrix $P$, under a specific policy, we let $\pi$ denote the stationary distribution.  The reward function is denoted with $r(x,u)$ where we assume that $|r(x,u)| \le \Gamma$. We consider a probabilistic policy mapping $\mu_\theta(u|x)$ which expresses the probability of the agent to choose an action $u\in U$ given that the agent is in state $x \in X$.

The \emph{Reward-To-Go} is a random variable that expresses the accumulative discounted rewards that the agent receives during its interaction with the environment
\begin{equation}
\label{eq:B_definition}
    B_\tau \triangleq \sum_{t=0}^{\tau} \gamma^t r(x_t,u_t),
\end{equation}
where $0 \le \gamma \le 1$ is the \emph{discount factor} and $\tau$ is the reward-to-go horizon. The horizon can be either finite, infinite, or stochastic. The goal of the agent is to find a policy that maximizes the so called \emph{Value Function}
\begin{equation}
    J(x) \triangleq \E_\theta\left[\left.B\right|x_0=x\right],
\end{equation}
where $\E_\theta [\cdot]$ is the expectation w.r.t. the MDP and the policy function that depends on $\theta$. For ease of exposition, we omit the subscript $\theta$ whenever it is clear from the context. Based on the reward-to-go, we define the risk measure to be
\begin{equation}
\label{eq:risk_definition}
    R(x)\triangleq \E \left[f\left(B - J(x)\right) |x_0=x \right],
\end{equation}
where $f(\cdot)$ is a function such as the square, absolute value, square root, etc. . Our objective in this case is 
\begin{equation}
\label{eq:original_constrained_optimization}
    \max_\theta J(x)\quad \textrm{s.t.} \quad R(x) \le D.
\end{equation}
Similarly to \cite{tamar2012policy}, we suggest to approximate this optimization problem with a soft constraint. We define
\begin{equation}
    \label{eq:rho_definition}
    \eta(x) = J(x)- \lambda g\left( R(x) - D \right), 
\end{equation}
where $g(\cdot)$ is a penalty function that is typically taken to be $g(x) = (\max (0, x))^2$, and $\lambda > 0$ is the penalty coefficient. Based on Eq.~\eqref{eq:rho_definition}, we have the  optimization problem 
\begin{equation}
\label{eq:soft_constrained_optimization}
    \max_\theta \left \{ \eta(x) \right\},
\end{equation}
whereas $\lambda$ increases, the solution of Eq.~\eqref{eq:original_constrained_optimization} converges to the solution of Eq.~\eqref{eq:soft_constrained_optimization}. We propose a solution to this problem that follows the  gradient descent approach by way of iterative updates to the value of $\theta$: 
\begin{equation}
\label{eq:PG_GD}
\theta_{t+1} = \theta_t + \alpha \nabla \eta(x).
\end{equation}

\subsection{Policy Gradient Methods}
Next, we recall the equations of the \emph{policy gradient} method  for the case of a finite time trajectory (see \cite{baxter2001infinite}). We can further generalize this method as follows. 

\begin{proposition}
Let $f: \mathbb{R}\rightarrow \mathbb{R}$ be a differentiable function. Then, 
\begin{equation}
\label{eq:gradient_expectation_B}
\nabla \E \left[\left.f(B) \right|x_0 \right] 
=
\E \left[ 
f\left(B \right)   
\sum_{t=0}^{\tau-1}\frac{\nabla P_\theta(x_{t+1}|x_t)}{P_\theta(x_{t+1}|x_t)}
\right],
\end{equation}
where the gradient $\nabla$ is taken w.r.t $\theta$.
\end{proposition}

\begin{proof}
\begin{equation*}
\label{eq:likeli1}
\begin{split}
\nabla_\theta& \E \left[\left.f(B) \right|x_0=x \right] \\
&\overset{a}{=}  \sum_{x_0^T} \nabla_\theta P_\theta(x_0^T)f\left(\sum_{t=0}^{T}\gamma^t r(x_t) \right) \\
&\overset{b}{=}  \sum_{x_0^T}P_\theta(x_0^T)\frac{\nabla_\theta P_\theta(x_0^T)}{P_\theta(x_0^T)} f\left(\sum_{t=0}^{T}\gamma^t r(x_t) \right). \\
\end{split}
\end{equation*}
where in (a) we changed order of the summation and the gradient, and in (b) we multiplied and divided by the same factor.

Next, for $\frac{\nabla P_\theta(z_0^T)}{P_\theta(z_0^T)}$ it is easy to show that
\begin{equation*}
\begin{split}
\frac{\nabla_\theta P_\theta(z_0^T)}{P_\theta(z_0^T)}
=
\sum_{t=0}^{T-1}\frac{\nabla P_\theta(z_{t+1}|z_t)}{P_\theta(z_{t+1}|z_t)}.
\end{split}    
\end{equation*}
Plugging this expression into \eqref{eq:likeli1} we get the desired result.
\end{proof}

Now, for general risk measures, we need to calculate the gradient $\nabla R(x)$ of Eq.~\eqref{eq:risk_definition}. Here, the inner part of the expectation (specifically, the term $J(\cdot)$)
also depends on $\theta$, which adds an additional step to our derivation:
\begin{equation}
\label{eq:grad_f_b_J}
\begin{split}
\nabla \E &\left[\left.f\left(B - J(x)\right) \right|x_0=x \right] \\
&= \nabla \sum_{x_0^\tau} P_\theta(x_0^\tau)f\left(B - J(x) \right) \\
&=  \sum_{x_0^\tau}\big[f\left(B - J(x) \right) \nabla P_\theta(x_0^\tau)\\
& \quad +  
P_\theta(x_0^\tau)\nabla f\left(B - J(x) \right)\big]. 
\end{split}
\end{equation}
Eq.~\eqref{eq:gradient_expectation_B} gives a simplified version of the first term in Eq.~\eqref{eq:grad_f_b_J}. As for the second term in Eq.~\eqref{eq:grad_f_b_J} we have
\begin{equation*}
\begin{split}
\nabla f\left(B -J(x) \right) 
=&
 - f' \left(B - J(x) \right) \nabla J(x).
\end{split}
\end{equation*}
To summarize
\begin{align}
\label{eq:grad_risk}
\begin{split}
&\nabla \E \left[\left.f\left(B - J(x)\right) \right|x \right] =\\
&  \sum_{x_0^\tau}\nabla P_\theta(x_0^\tau)\big[f\left(B - J(x) \right)
 - f' \left(B - J(x) \right) \nabla J(x)\big].
\end{split}
\end{align}

\section{The Grid World Setup}
\begin{figure}
    \centering
    \includegraphics[scale=0.55]{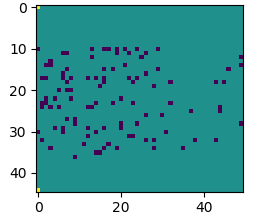}
    \caption{An example of the grid world. The starting point is located in the upper left corner whereas the end point is in the lower left corner, both highlighted in yellow. The mines are marked by the dark-blue points.}
    \label{fig:GridWorldMineExample.png}
\end{figure}

Although our framework is  quite general, we present our risk-shaping approach (in the following section) and focus our experiments on the well-studied Grid World setting~\cite{sutton2018reinforcement}.
 There, the objective is to find a path from a prescribed starting point to a designated target point. Each square represents a location on a two dimensional grid. Within this simple two-dimensional space, and between the start and termination points, there are two special types of entries. The first one are \emph{mines}, which are associated with a negative reward\footnote{For the simplicity of our discussion we consider uniform reward values, but this restriction can be easily lifted.} of $r_{mine}<0$ that is given to the agent whenever it steps on them. The probability that a location will contain a mine is linear and monotonically decreasing with the the $x$ coordinate, where $0 \le x \le x_{max}$):
\begin{equation*}
    \textrm{Prob(location $x,y$ is mine)} = p_{mine} \frac{x_{max}-x}{x_{max}},
\end{equation*}
Typically, we set $p_{mine}=0.2$. The second special type of entry is an entry with an associated reward of $r_{target}$. All other entries, are associated with zero rewards. At every time step, the agent can move \emph{up, down, right}, or \emph{left} to any of its adjacent squares, and we let $U$ denote the set of these four directions/actions. An illustration of this grid world is given in Figure \ref{fig:GridWorldMineExample.png}. Additionally, the agent's movement is subject to control noise that, with probability $p_{noise}$, causes the agent to move in a uniformly random direction in $U$. 

\section{Risk-Shaping and General Risk Functions}
\label{sec:risk_shaping}
In this section we demonstrate that risk functions do not necessarily need to be \emph{coherent}, and in particular, can be non-convex. We illustrate this with particular instance of the \emph{gambler's ruin problem} \cite{norris1998markov}. Many real life problems, and in particular problems in economics and finance, lie well within this domain of problems \cite{duff1974gamblers,wilcox1976gambler}.

Our specific variant of the gambler's ruin problem relies on a Markov Reward Process (MRP; \cite{howard2012dynamic}), and is a more powerful setup than the Markov chain model. The state space is infinite and the states are denoted by $\mathbb{Z} = 0,1,2,\ldots$. These state values denote the amount of money (in dollars) in the current possession of the agent. The process terminates when the agent reached the state $\$0$, which indicates the bankruptcy of the agent. At each time step, the agent gambles, and wins a dollar with probability $\frac{1}{2}$, and otherwise it loses a dollar (with probability $\frac{1}{2}$). The agent can only gamble as long as its fortune is strictly positive (we exclude borrowing money). We define the following condition for risk: $p_k(m)$ is the probability that the agent will go bankrupt within $k$ steps, given an initial fortune of $\$m$. Therefore, the risk in this case is a probability; i.e., $R(x=m) = p_k(m)$, where $k$ is the \emph{risk-look-ahead} parameter. In other words, $k$ dictates the interval upon which we optimize our risk.

We want to solve Eq.~\eqref{eq:risk_definition} according to this model. We know the risk for each state $x$ and we can calculate $J$ using Monte Carlo simulation or any other Policy Evaluation (PE; \cite{bertsekas2005dynamic}) and has a realization of the $f(\cdot)$ argument. Based on that, we get samples for the risk as function of the state $x$. Now, we are interested in learning a function $f$ that best fits the model. We refer to this process of learning this function $f$ from the data as \emph{Risk Shaping}.

\begin{figure}
    \centering
    \includegraphics[scale=0.4]{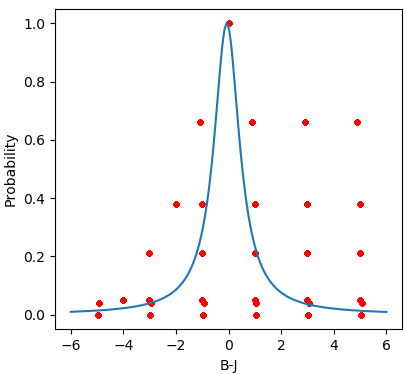}
    \caption{Samples of a Monte Carlo experiment as described in Section \ref{sec:risk_shaping}. Each sample in the graph (red dots) is a relation between a realization of $B-J$ and a probability $p_k(m)$. The continuous blue line describes the fit to the function $f$.}
    \label{fig:risk_shaping}
\end{figure}

We ran the process described above and obtained samples that describe the correspondence between $B-J$ and $R(x)$, as described in Eq.~\eqref{eq:risk_definition}. This correspondence is depicted in Figure \ref{fig:risk_shaping} (red dots). We selected as a model for these samples the equation $1/(1+b*(x-c)^2)$, where $b >0$ and $c\in \mathbb{R}$.

We remark that as opposed to common practice, in this particular example, limiting the variance (as in \cite{tamar2012policy,tamar2013temporal} and its followup work), or optimizing the VaR or CVaR is arguably disadvantageous. The high-risk regime in this example is characterized by low variance (due to the vicinity to $x=0$). Moreover, for $x>k$ the variance is maximal and the risk is zero. This is the reason for the non-convexity of the $f(\cdot)$, as illustrate Figure~\ref{fig:risk_shaping}.

\section{Neural Architectures}
\label{sec:neural_architectures}
In this section we describe a neural architecture for estimating the different components. We propose the Actor Risk-Critic Value-Critic Architecture (ARCVC), which consists of three main components: (1) An actor for the policy (2) a value-function critic, and (3) a risk-function critic. The value function critic component is a standard (and not necessarily linear) function approximation for $J(\cdot)$, denoted by $\hat{J}(x,\omega)$. We focus in this section on the other two components. We recall that, in order to estimate the value function, most techniques (excluding Monte-Carlo simulations; \cite{sutton2018reinforcement}) use some form of the Bellman equation. For some concrete cases of risk functions, a similar approach may be applied, such as in the variance case \cite{sobel1982variance}. However, as noted by the authors, such closed form equations exist only for a limited cases. To address this difficulty, we employ Monte Carlo simulations  in the proposed architectures.

\begin{figure}
    \centering
    \includegraphics[scale=0.44]{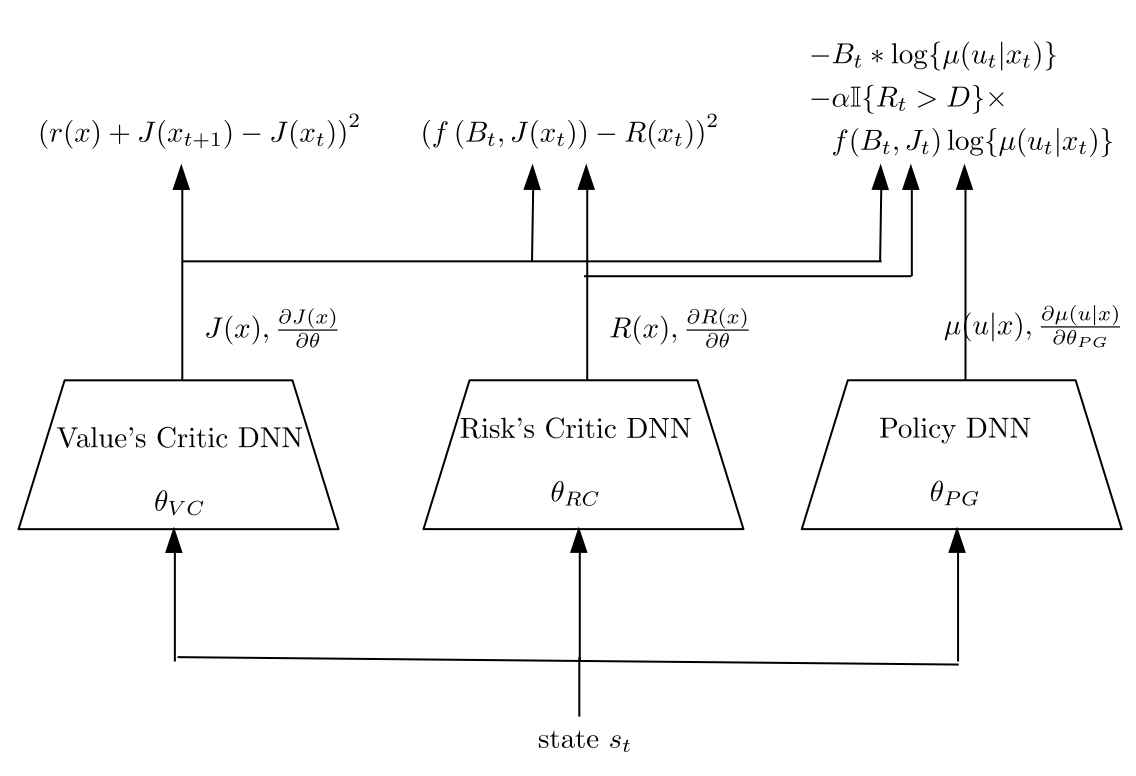}
    \caption{The risk critic architecture for mitigating risk in MDPs. The left most network function is to estimate the value function (value function critic). The middle network role it to estimate the risk of the current state (risk critic). The right most network is the action network, that carries out a policy gradient for the policy. We denote the parameters of the value function network with $\theta_{VC}$, the parameters of the risk network with $\theta_{RC}$, and the parameters of the policy gradient network with $\theta_{PG}$.}
    \label{fig:RiskArchFull.png}
\end{figure}

We propose the use of a finite time buffer, denoted by FTB, to collect the $\tau$ recent samples of the reward, and, once  collected, we use them to compute an estimate of the risk according to Eq.~\eqref{eq:risk_definition}:
\begin{equation}
\label{eq:risk_rho_t}
   \rho_t \triangleq f\left(B_\tau(x_t) -\hat{J}(x_t)\right),
\end{equation}
Such a buffer can be implemented by using a queue. The running time complexity and space complexity of such architecture using a queue are  $\mathcal{O}(1)$ and $\mathcal{O}(\tau)$, respectively, for computing the risk sample. Therefore, the \emph{loss function} that the Risk-Critic is minimizing is
\begin{equation}
\label{eq:risk_loss}
    loss^{\textrm{risk}}_t \triangleq L\left(\hat{R}(x)- \rho(x) \right).
\end{equation}
A schematic illustration of this Architecture is provided in Figure \ref{fig:RiskArchFull.png}. We can see that three networks are involved: one for the actor, one for the value function critic, and one for the risk critic. We can see that the value function is needed by the risk network: it is used as a  \emph{reference} value for computing the risk. The roles of the risk network itself, w.r.t. the policy network, are twofold. First, it provides indication of whether or not the risk constraint is violated. The second role pertains to the objective itself: whenever we violate the constraint it adds to the general objective function and pulls the policy gradient towards the direction that minimizes the risk value.

\subsection{A Compact Architecture}
\label{subsec:compact_arch}
As was shown in the previous section, a naive architecture involves three networks. In this section, we show how to reduce the network size making several modifications. This reduces the computational overhead, both in terms of the running time complexity as well as the space complexity.

\subsubsection{Changing the Reference}
\label{subsec:changing_reference}
We suggest a more compact architecture that does not involve the value function network. First, consider a slightly more general version of the  objective function:
\begin{equation*}
    \eta(x) = \E \left [B +  \alpha f(B-\nu(x)) g \{E[f(B-\nu(x))] - D \} \right].
\end{equation*}
Setting $\nu(x)=J(x)$ admits the original setup presented in the previous section (see Eq.~\eqref{eq:risk_definition}). We substitute the  soft constraint with the following 
\begin{equation*}
\begin{split}
    \hat \eta(x) =& \E \big [B +  \alpha (f(B-\nu(x))-D)\\
    &\times \E[ g \{f(B-\nu(x)) -D) \}] \big],
\end{split}
\end{equation*}
where the argument of the $g$ function and the constraint are identical and where we denoted this by $\hat \eta (\cdot)$. 
We consider several ways to set $\nu(x)$.

\begin{enumerate}
    \item \textbf{Changing reference $J(x)$ to $\eta(x)$.} We propose to replace $J(x)$ with $\eta(x)$ in Eq.~\eqref{eq:rho_definition}. Therefore, the objective is
\begin{equation}
\label{eq:bootstrapped_objective}
\begin{split}
\hat \eta(x) 
=& \E_\theta \big [B +  \alpha (f(B-\eta(x))-D) \\ 
 &\times g \{f(B-\eta(x)) -D) \} \big].
\end{split}
\end{equation}
The meaning of Eq.~\eqref{eq:bootstrapped_objective} is the following: instead of just measuring the average distance from the mean trajectory, as captured by the subtraction of $J(\cdot)$ from the trajectory, we also \emph{include the soft constraint} in the distance function. This modified version has an interesting property: It is easy to show that when the constraint is satisfied, the function converges to $J(\cdot)$. In other words, we get a similar objective for satisfying the constraints, i.e.,
\begin{equation*}
    \lim_{f(B-\eta) \rightarrow D} \hat \eta(x) = J(x).
\end{equation*}
The downside of this replacement is that it does not reduce the complexity. Now, instead of having to estimate $J(\cdot)$ we only need to estimate $\eta(\cdot)$.
\item \textbf{Changing reference $J(x)$ to $\E[J(x)]$.} We propose to replace $J(x)$ with $\bar J\triangleq \E_{x \sim \mathbf{\pi}} [J(x)]$ where $\mathbf{\pi}$ is the stationary distribution. In order to estimate $\bar J$ we propose the following Stochastic Approximation iteration~\cite{kushner2003stochastic}:
\begin{equation*}
    \bar J_{t+1} = \bar J_{t} + \alpha_t \left(r(x_{t+1},u_{t+1}) - \bar J_{t}  \right),
\end{equation*}
where $\alpha_t$ is the step of the iteration that may be a ``small" constant or decreasing time step that behaves like $\mathcal{O}(\frac{1}{t^{\delta}})$, for $\frac{1}{2} \leq \delta \leq 1$. The advantage of this iteration is clear: instead of maintaining a network for $J(\cdot)$, we only need to perform scalar updates. The downside to this approach is the potential loss in the accuracy of the risk estimate. However, in many cases, this approximation proves to be relatively good.
\item \textbf{Changing reference $J(x)$ to constant.} Another possibility would be to replace $J(x)$ with some constant, based on some prior information we obtain from a domain expert. We do not study this approach in the present work, although it might prove to be stable and eventually beneficial. 
\end{enumerate}

\subsubsection{Sample Based Penalty Function}
\label{subsec:SampleBasedPenaltyFunction}
Another simplification is to remove the dependency of the penalty function on $R(x)$. The risk network role is map from state to the risk associated with this state. We propose to base the penalty function solely on a sample that represents the current risk. Similarly to Section \ref{subsec:changing_reference}, we can shrink the architecture by a whole estimation network. Later, in the experiments, we demonstrate that such change does not reduce the agent performance, neither in the accumulated reward, nor in the risk quality. The basic algorithm for the architecture AVCRC (in episodic form) is summarized in Algorithm \ref{alg:ARCVC}.
\begin{algorithm}[tb]
   \caption{Episodic ARCVC Algorithm}
   \label{alg:ARCVC}
\begin{algorithmic}
   \STATE {\bfseries Input:} 
   \STATE {\bfseries Set parameters:} $\gamma$, $\lambda$, $D$, $f(\cdot)$, $g(\cdot)$, $\alpha$.
   \STATE {\bfseries Choose Reference Method:} $rm$ - to be state based or global based or constant  (Section \ref{subsec:compact_arch})
   \STATE {\bfseries Choose Penalty Method:} $pm$ - to be Sample Based or Risk Network Based (Section \ref{subsec:sample_based_penalty})
   \STATE {\bfseries Initialize:} Initialize FTB (Finite Time Buffer)
   \STATE {\bfseries Initialize:} PG neural network (actor) with parameter $\theta_{PG}$.
   \IF{$rm$ is state based}
   \STATE {\bfseries Initialize:} Value Network (critic) with parameter $\theta_{VC}$
   \STATE {\bfseries Initialize:} Replay Buffer 
   \ENDIF
   \IF{$pm$ is Risk Network Based}
   \STATE {\bfseries Initialize:} Risk Network (critic) with parameter $\theta_{RC}$
   \ENDIF
   \FOR{each episode}
   \STATE {\bfseries Empty} FTB
   \STATE {\bfseries Collect} state $x_t$, reward $r_t$, action $a_t$ for length-$\tau$ batch.
   \STATE {\bfseries Construct} $B_{\tau}$ based on $r_t$ from FTB according Eq.~\ref{eq:B_definition}
   \STATE {\bfseries Push} into Replay Buffer $(x_t, r_t, a_t, x_{t+1}, B_t)_{t=1}^\tau$
   \STATE {\bfseries Optimize} PG according to Eqs.~\eqref{eq:rho_definition},~\eqref{eq:PG_GD},~\eqref{eq:grad_risk} and according to Reference Method and Penalty Method.
   \STATE {\bfseries Optimize} Value Network according to DQN algorithm \cite{mnih2015human}.
   \STATE {\bfseries Optimize} Risk Network according to Eq.~\eqref{eq:risk_rho_t} and Eq.~\eqref{eq:risk_loss}.
   \ENDFOR
\end{algorithmic}
\end{algorithm}

\section{Experiments}
\label{sec:experiments}

In this section empirically investigate the behavior of different risk measures and strengthen our understanding of the specific trade-offs of such scenarios. 

In all experiments, we set $\gamma=0.9$. The policy gradient network has two layers with a ReLU \cite{goodfellow2016deep} activation between them and a soft-max output layer. Both critics, the value estimation network and the risk estimation network have three layers where the first two activation functions layers are ReLU, and the output layer is linear. For optimization we use the Adam optimizer \cite{kingma2014adam}, which gave us the best and most stable results. 

\subsection{The Risk Violation Rate - A Measure for Examining the Risk}
In order to define the efficacy of our risk is in practical terms, we need to quantify how well it manages to reach its objective while minimizing the likelihood of violating its prescribed risk constraint. We propose the following method for grading.

\begin{definition}
\emph{Risk Violation Rate} is the fraction of times in which constrained Risk problem violated the risk constraint.
\end{definition}
In the experiments we describe below, we use this measure in order to estimate how good a method is for risk estimation and satisfying the risk constraints. Additionally, we couple the risk violation rate with the algorithm's success rate, as otherwise the algorithm could trivially uphold the constraint without actually reaching the original objective.\footnote{This is analogous to the precision-recall trade-off \cite{powers2011evaluation}}.

\subsection{Comparison of Different Risks Measures}
We examined our algorithm on three representative risk functions: (1) ``One Sided Variance" (denoted with $Var^-$) given by $f(B-J) \triangleq (B-J)^2 \mathbb{I}\{B-J \le 0\}$, (2) ``One Sided Absolute Value" (denoted by $Abs^-$) given by $f(B-J) \triangleq |B-J| \mathbb{I}\{B-J \le 0\}$, and (3) ``One Sided Square Root" (denoted by $Sqrt^-$) that is given by $f(B-J) \triangleq |B-J|^{\frac{1}{2}} \mathbb{I} \{ B-J \le 0 \}$. The reason for taking one-sided  functions is that we are only concerned about \textit{negative rewards}.

Now, suppose we want to compare these risk functions. For appropriate comparison, it is easy to show that a constraint value $D$ should scale differently for different risk functions. Indeed, the constraint value $D$ is scaled by a constant, as dictated by the  function $f(\cdot)$ in hand. For example, if we use the One Sided Absolute Value function with a constraint value $D$, then its analogous constraint value for the One Sided Square Root will be $\sqrt{D}$, and similarly, $D^2$ for the One Sided Variance function.

\begin{figure}
    \centering
    \includegraphics[scale=0.5]{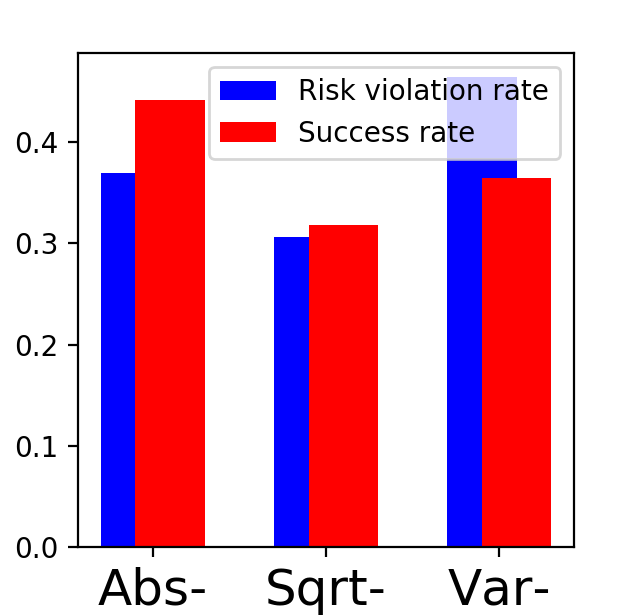}
    \caption{Risk Violation Rate and Success Rate for 3 different risk measures. Refer the text for more details.}
    \label{fig:risk_violation_vs_success}
\end{figure}

In our experiments, we set the constraint parameter $D=0.1$. This value turned out to push the algorithm to display quite an interesting behavior. Indeed, it  caused the algorithm to have difficulty with balancing both the risk satisfaction constraint and the reward-to-go maximization. The results are depicted in Figure \ref{fig:risk_violation_vs_success}. We can see that the non-coherent risk measures (the $Sqrt^-$ function) may be beneficial in some cases. First, they are aggressive in forcing the algorithm not to violate the constraint. Second, since the risk constraint is enforced during training, it can be regarded as \textit{safe exploration}. On the other hand, as evident the results that overly aggressive risk functions can deteriorate the success rate. As pointed-out by \cite{tamar2012policy}, dealing with risk in the context of RL can sometimes pose a trade-off between performance and the risk constraint satisfaction.\footnote{The rest of the parameters for this experiments are: grid world of size $20\times25$, D=0.1, $\lambda=10$, batch size of 100, 1000 episodes each run, repeated 50 times for each risk function, $\gamma=0.9$, ADAM optimizer, PyTorch ver. 1.0.0, and learning rate $0.001$ for all networks.}.

\subsection{Changing the Reference}
In the next experiment we examine the effect of changing the reference of the risk, as described in Section \ref{subsec:changing_reference}. We compared the original risk definition (Eq.~\ref{eq:risk_definition}) to the global reference $\E[J(x)]$. We define a score function that measures how much the risk based on a global reference (i.e., 
$\epsilon_G\triangleq\E[f(B(x) - \E_\pi[J(x)])]$) 
deviates from the original risk weighted by the stationary distribution (i.e., 
$\epsilon_\pi\triangleq\E_\pi[f(B(x) - J(x)]$).

\begin{figure}
    \centering
    \includegraphics[scale=0.45]{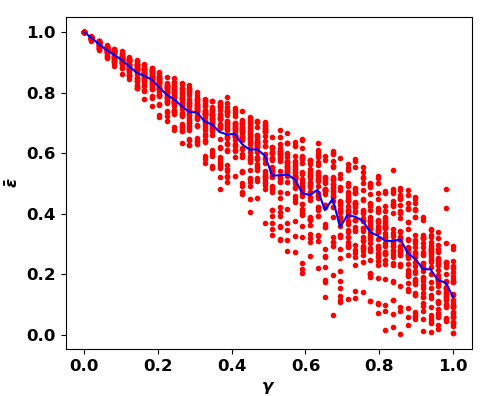}
    \caption{Plot of $\bar \epsilon$ vs. the discount factor $\gamma$. In red dots different repeats are presented. The blue line is the average of $\bar \epsilon$ for each $\gamma$. This plot is based on 50 values of $\gamma$ and $30$ runs for each value.}
    \label{fig:epsilon_bar}
\end{figure}

In order to express the deviation as a score, we define the distance between $\epsilon_G$ and $\epsilon_\pi$ to be
\begin{equation*}
    \bar \epsilon = \frac{|\epsilon_G - \epsilon_\pi|}{\max(\epsilon_G , \epsilon_\pi)}.
\end{equation*}
Figure \ref{fig:epsilon_bar} depicts the empirical distribution of $\overline{\epsilon}$ as we vary the discount factor $\gamma$ between $0$ and $1$. We see that the difference between the global reference and the original references is diminishes as the discount factor approaches $1$. 

\subsection{Sample Based Penalty Function Estimation}
\label{subsec:sample_based_penalty}
The main objective of the risk network is to faithfully capture the risk for the penalty function $g(\cdot)$. In this experiment we show that we do not lose much when we replace the risk network signal in the penalty function of Eq.~\ref{eq:rho_definition} with a single sample based estimation. We conducted an experiment on the grid world environment, with $50$ different mine layouts, to estimate the penalty function based on the risk network output. We estimate the penalty function for the same 50 mine layouts, based on a single sample at each time step. More specifically, if we examine the penalty function in Eq.~\ref{eq:rho_definition}, we replace the term $g(R(x)-D)$ with the term $g(f(B-J(x))-D)$. 

\begin{figure}
    \centering
    \includegraphics[scale=0.4]{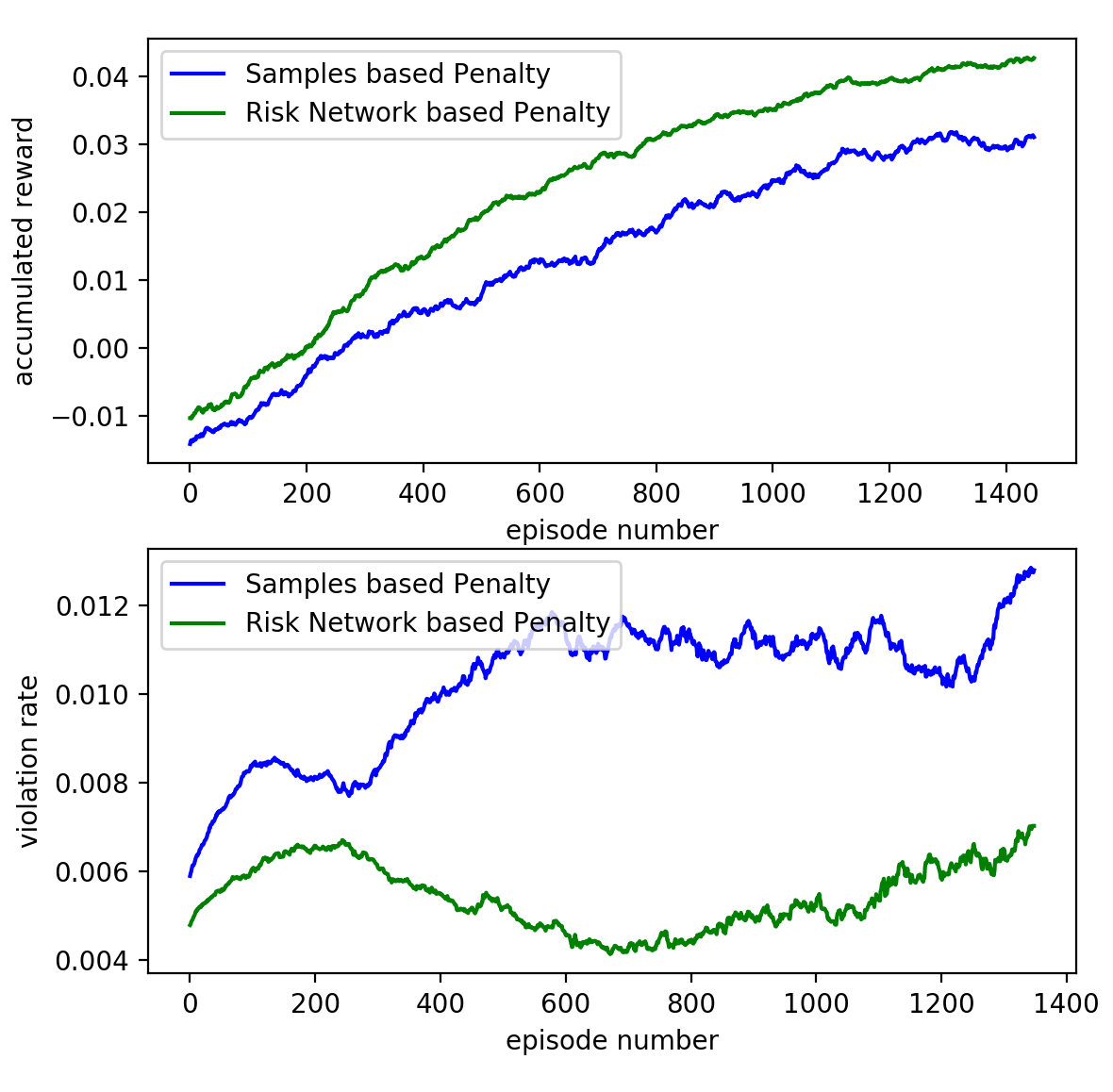}
    \caption{A comparison between Samples Based Penalty and Risk Network Based Penalty. The upper plot depicts the comparison of the accumulated rewards for both methods and in the lower plot depicted a comparison for the violation rate for both methods.}
    \label{fig:sample_based_vs_v_net}
\end{figure}

The results are depicted in Figure \ref{fig:sample_based_vs_v_net}. On one hand, there is a clear deterioration in terms of both the average accumulated reward, as well as the risk violation. On the other hand, this modification to the architecture eliminated the need for training an additional network (the risk network), and therefore results in massive savings in both memory and running time.

To Summarize, one can get a compact architecture that is based only on the policy gradient network if (1) the value function network is replaced with a global reference and (2) the risk function is replaced with a single sample estimator.

\section{Conclusions and Future Work}
\label{sec:conclusions}
We have shown that natural risk measures that are extracted from some simple domains do not exhibit necessarily some of the coherent risk requirements. In addition, we described a procedure for extracting an appropriate risk function from data,  enabling the domain expert to understand and then approximate risk functions. Being able to shape the risk of a given problem, and tailor it to problem specifics is important because risk is more difficult to understand (and design) than the reward and consequently the value function.

We believe that investigating methods and best practices in risk shaping for different domains is paramount for the applicability of risk awareness in planning, RL problems, and MDPs in general.

In this work we did not apply a holistic approach, i.e, for  given realizations of the agent interacting with an environment, we provided a method for extracting the risk function (i.e., risk shaping) and afterward, we showed a method that can apply this general risk function in the MDP (i.e., applying generalized risk functions). As a future direction, we propose to interleave these two important methods into a single holistic algorithm where using light supervision we learn the risk \emph{while} solving the MDP. Also, in the context of safe exploration, we suggest to incorporate constraining the violations rate as well.

\bibliographystyle{named}
\bibliography{bib}

\end{document}